\documentclass{article}

\usepackage{arxiv}

\usepackage{subfigure}
\usepackage[utf8]{inputenc} 
\usepackage[T1]{fontenc}    
\usepackage{hyperref}       
\usepackage{url}            
\usepackage{booktabs}       
\usepackage{amsfonts}       
\usepackage{nicefrac}       
\usepackage{microtype}      
\usepackage{lipsum}
\usepackage{graphicx}

\title{Autotm 2.0: Automatic Topic Modeling Framework for Documents Analysis}

\author{
 Maria Khodorchenko \\
  ITMO University \\
  Saint-Petersburg, Russia \\
  \texttt{mariyaxod@yandex.ru} \\
   \And
 Nikolay Butakov \\
  ITMO University \\
  Saint-Petersburg, Russia \\
  \texttt{alipoov.nb@gmail.com} \\
  \And
 Maxim Zuev \\
  ITMO University \\
  Saint-Petersburg, Russia \\
  \texttt{zuev.m67@gmail.com} \\
    \And
 Denis Nasonov \\
  ITMO University \\
  Saint-Petersburg, Russia \\
  \texttt{denis.nasonov@gmail.com} \\
}

\begin{document}
\maketitle

\begin{abstract}
In this work, we present an AutoTM 2.0 framework for optimizing additively regularized topic models. Comparing to the previous version, this version includes such valuable improvements as novel optimization pipeline, LLM-based quality metrics and distributed mode.
  
  AutoTM 2.0 is a comfort tool for specialists as well as non-specialists to work with text documents to conduct exploratory data analysis or to perform clustering task on interpretable set of features. Quality evaluation is based on specially developed metrics such as coherence and gpt-4-based approaches. Researchers and practitioners can easily integrate new optimization algorithms and adapt novel metrics to enhance modeling quality and extend their experiments.

We show that AutoTM 2.0 achieves better performance compared to the previous AutoTM by providing results on 5 datasets with different features and in two different languages.
\end{abstract}


\section{Introduction}
Topic modeling is a well-known technique for modeling the internal structure of a text corpora, represented as a set of interrelated word sets known as topics. Starting from Latent Semantic Allocation (LSA) \cite{deerwester-indexing-1990} and Non-negative Matrix Factorization (NMF) \cite{10.5555/3008751.3008829} to probabilistic and neural approach, topic modeling proved to be a valuable tool to solve a range of practical tasks \cite{INR-030,ABDELRAZEK2023102131}. One of the key features of topic modeling lays in the interpretability of resulting representations, that enables easier comprehension of complex datasets and helps in meaningful insights extraction.

To be useful, topic models should be flexible enough to model various corpora of different nature, origin, and language. Which requires the model to be carefully tuned for the corpora in consideration at the moment, and usually is closely connected with the amount of hyperparameters the model has. This is especially true for additively regularized topic models that represent semi-probabilistic group of methods revealing great adaptability, but requiring setting a high number of parameters and expertise to do that properly.  

This paper presents AutoTM 2.0 framework that allow effective usage of additively regularized models, as they provide the most flexible way to process datasets with different statistical characteristics.

Our main contributions can be summarized as follows:
\begin{itemize}
    \item significant simplification of the use of flexible additively regularized models by offering automatic single-objective optimization procedures.
    \item Offering metrics that closely align with human judgment.
    \item  Enabling cost-effective inference and rapid training for extensive text corpora.
    \item Providing a Python library with distributed capabilities to conduct extensive experiments or manage large datasets.
\end{itemize}

\section{Related work}
All the topic modeling approaches can be roughly divided into four groups: statistical (LSA, NMF), probabilistic (pLSA, LDA), semi-probabilistic (ARTM \cite{Vorontsov2014AdditiveRF}) and neural (BERTopic, ETM). All of them have their pros and cons. The first 3 groups have quick inference speed and require few computational resources in comparison to the group 4.  Methods of group 4 may better capture semantics of the texts and corpora, but can be limited by size of corpora. While it is possible to process huge datasets with large language models, it is still an expensive task in terms or cost and computational resources. Interpretability of model parameters and handling sparsity are not its strong sides too \cite{ABDELRAZEK2023102131}. Among the first 3 groups, statistical and probabilistic methods usually have limited number of hyperparameters that require careful choice or tuning. Though methods of both these groups are less flexible and may struggle to properly model complex internal structure of the corpora. ARTM approach is much more flexible due to various regularizers that can be applied during training to model various peculiarities of a specific dataset, but it may require to tune much higher number of hyperparameters and even the sequence in which particular regularizers to be applied.       

There exist several frameworks on topic modeling that work with different types of models. Gensim library\footnote{https://github.com/piskvorky/gensim} is one of the most known topic modeling libraries and include several classical topic models. MALLET \cite{McCallumMALLET} provides classic probabilistic topic models. BERTopic\footnote{https://github.com/MaartenGr/BERTopic} \cite{grootendorst2022bertopic} is an open-source library that incorporates a set of contextualized methods. Language models are used as a source of pretrained vector emebbings for further. OCTIS \cite{terragni-etal-2021-octis} is a framework that implements a range of topic models such as classical variants from gensim library (LDA, LSI, etc.).

However, frameworks that work with ARTM models are much less presented. This method was first implemented in the BigARTM C++ library \cite{10.1007/978-3-319-26123-2_36} and later made more available for users via Python wrappers and TopicNet library \cite{bulatov-etal-2020-topicnet}. The latter simplified working with BigARTM by automatically setting some of parameters but still demanded a deep understanding of regularizers and their specifics to build a solid training pipeline. 

First version of AutoTM \cite{Khodorchenko2023-hy} framework proposed introduced a genetic algorithm that search solution in the form of a fixed-sized vector representing strategy of ARTM training in several sequential steps with different regularizers. The proposed algorithm showed its efficiency in comparison with other approaches \cite{10.1093/jigpal/jzac019,9953874,khodorchenko2020optimization}.  Thus, AutoTM 1.0 proposed a way to improve qulity of ARTM models without requiring much expertise from the end user. 

The new version we propose in this paper takes it further by introducing a new form of solution (dynamic vector) with varying order of regularizers being applied that significantly improves the quality and a new quality measure based on llm. Thus, we present a framework specially focusing on ARTM models that greatly simplifies the use of the approach without hindering its flexibility and making it easier and more efficient to be applied for practical use cases.

\section{Framework design}

\begin{figure}[h]
  \centering
  \includegraphics[width=0.8\linewidth]{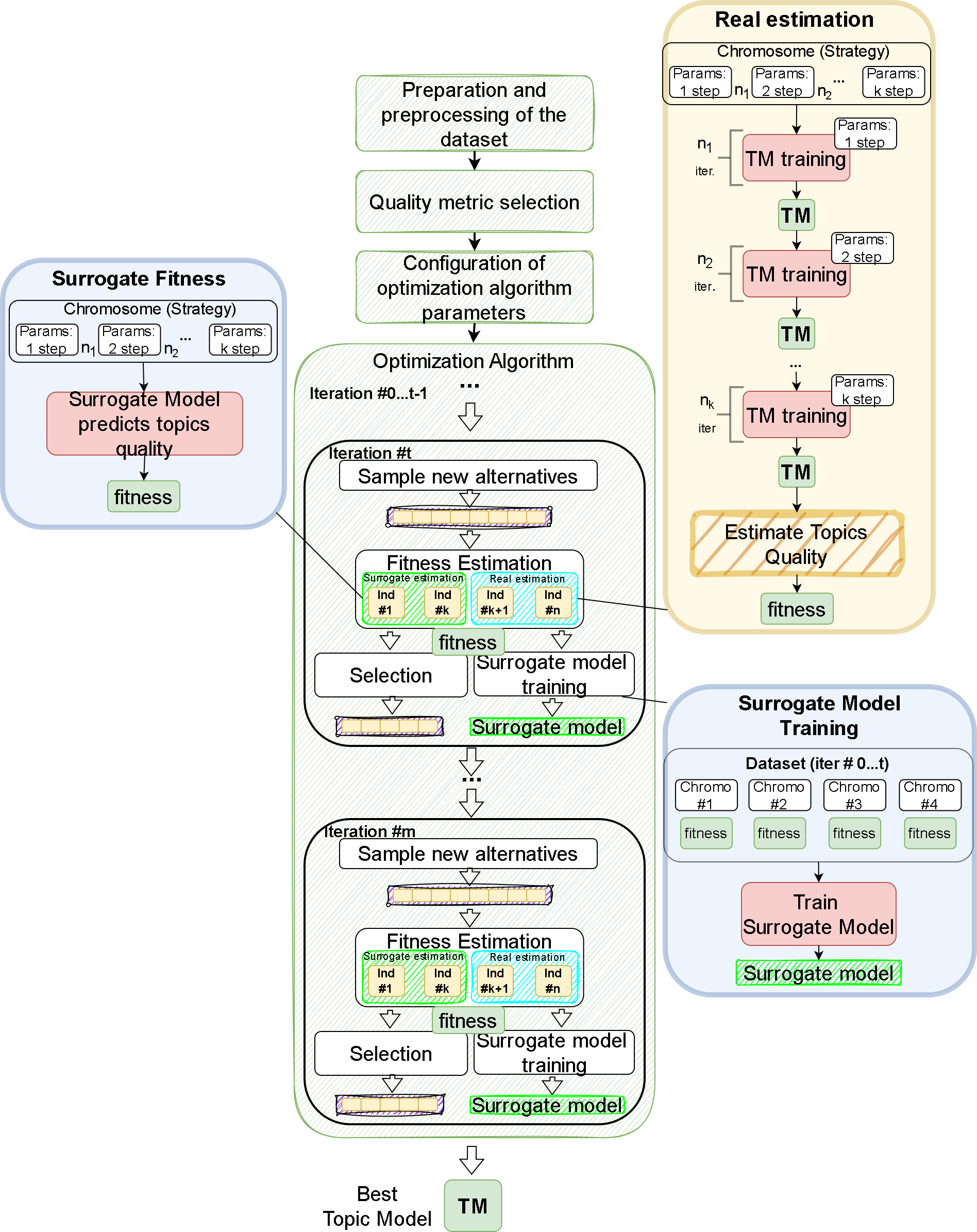}
  \caption{General design of the AutoTM 2.0 framework.}
  \label{fig:arch}
  
\end{figure}

General AutoTM pipeline is depicted on Fig.~\ref{fig:arch} and starts with the preparations of the dataset. All the preprocessing procedures are incorporated in the framework and described in section 3.1. 

A basic optimization pipeline on prepared data consists of several optimization iterations where a set of new models are trained, and, depending on the surrogate option flag, can be calculated as is or substituted by a surrogate approximation. 
It is important to note that optimization results highly depend on the metric that can be selected from available in the framework metrics.
The training procedure itself is stage-based and starts with BigARTM model initialization and model is tuned step-by-step on a specific set of parameters.

After the predefined amount of iterations (or on early stop trigger) AutoTM returns the best model and a set of topics that can be used to solve the domain tasks.

The framework is available as a Python library and provides a command-line interface. Details on distributed variant for large text corpora is provided in section 3.4.

\subsection{Dataset preprocessing}

AutoTM does not require user to make special data preparation and incorporates several steps for dataset preprocessing, including html tags cleaning, lemmatization, stop-word removal and cooccurrance vocabulary preparation, ppmi calculations and batching as a special step to convert the data to underlying BigARTM-compatible format.

It should be noticed that AutoTM implements special lemmatization and stopwords for English and Russian. At the same time, other languages may require additional preprocessing steps (and special lemmatization) to improve modeling quality.

\subsection{Optimization Approaches}

\subsubsection{Optimization pipelines}

AutoTM already contained an implementation of \cite{10.1093/jigpal/jzac019} with fixed-size parameter representation (will be mostly referred as "fixed-size" solution during the paper). AutoTM 2.0 in its order introduce a novel and more flexible parameter model (referred as a "graph-based pipeline") that allows customization of the number and types of stages, similarly to the TPOT pipeline optimization method \cite{10.1093/bioinformatics/btz470}.

New parameter representation is designed as a pipeline consisting of multiple stages, each representing a distinct training cycle of the ARTM model. Within each stage, various parameters can be specified, such as the number of passes through the dataset, the type of regularizer, and the specific parameters associated with that regularizer. 

\begin{figure}[h]
  \centering
  \includegraphics[width=0.8\linewidth]{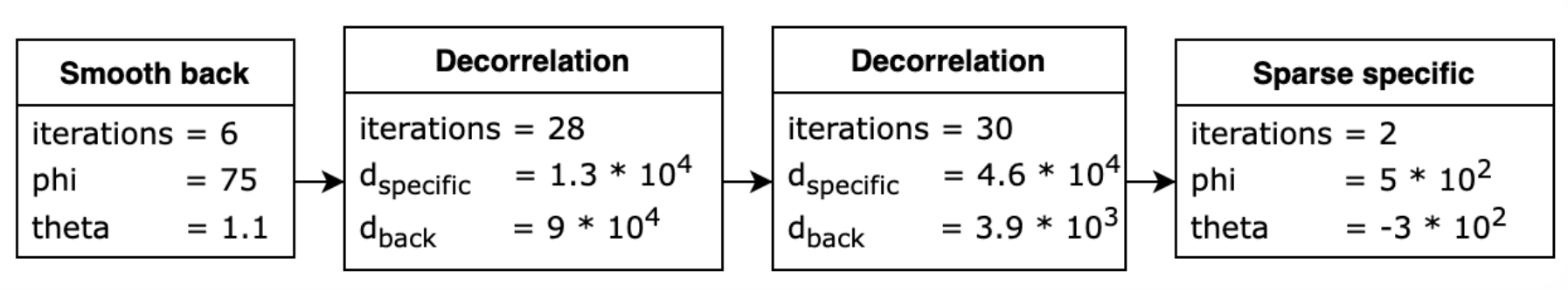}
  \caption{An example of a pipeline with 4 training stages. }
  \label{fig:pipeline}
  
\end{figure}

An example of a pipeline with 4 training stages is presented in Fig.~\ref{fig:pipeline}. In the first training stage of 6 iterations, a smoothing regularizer for background topics is enabled with parameters $\phi$ and $\theta$. In the
further 28 iterations, a decorrelation regularizer is introduced to the pipeline. Moving on to the third training stage, which spans across 30 iterations, the
pipeline resets the decorrelation parameters. The final training stage consists
of 2 iterations and incorporates a sparsing of the specific topic's regularizer.

We have introduced a number of new mutations to enhance the functionality of the graph-based model. These mutations provide more options for manipulating the stages within the pipeline. The available mutations include: adding a new random stage, removing a specific stage, swapping the positions of two stages, and mutating a random stage. When performing a crossover of two pipelines, a random crossover point is chosen in each pipeline. At this point, the heads and tails of the pipelines are swapped to create new pipelines.

The framework provides two basic algorithms for ARTM hyperparameters tuning, namely Bayesian optimization (BO) and the genetic algorithm (GA). While BO can be applied for only fixed-size solution, GA has appropriate operators for both variants.

\subsubsection{Quality estimation} Except from the coherence-based metric from \cite{10.1093/jigpal/jzac019} that has a high correlation with human judgement and implementations of the basic automatic metrics for quality evaluation (such as NPMI \cite{10.1145/1816123.1816156}, switchP \cite{lund-etal-2019-automatic}), there is also a variant of LLM-based metric inspired by \cite{stammbach-etal-2023-revisiting}.  

Thus, we present a default template for  gpt4o-based metric with the following prompt that aligns with the evaluation technique described in \cite{9953874}:
\textit{'''You are a helpful assistant evaluating the top words of a topic model output for a given topic. 
Please rate how related the following words are to each other on a scale from 1 to 4 ("1" = poorly related, "2" = rather poorly related, "3" = rather related, "4" = very related). 
Reply with a single number, indicating the overall appropriateness of the topic.'''} Comparison with the other evaluation approaches are presented in section 4.3. It is also possible for a user to change the default prompt and model to better satisfy his or her needs.

\subsection{Surrogate modeling}

Surrogate modeling in AutoTM \cite{10.1093/jigpal/jzac019} is a method where a surrogate function is created based on fitness evaluations for trained topic models on previous iterations. This surrogate function is based on a simple ML model (such as random forest regressor) and serves as a proxy for the actual computationally intensive calculations replacing them with a prediction and thus reducing the amount of heavy computation required while still maintaining high-quality results.

\begin{figure}[h!]
  \centering
  \includegraphics[width=0.8\linewidth]{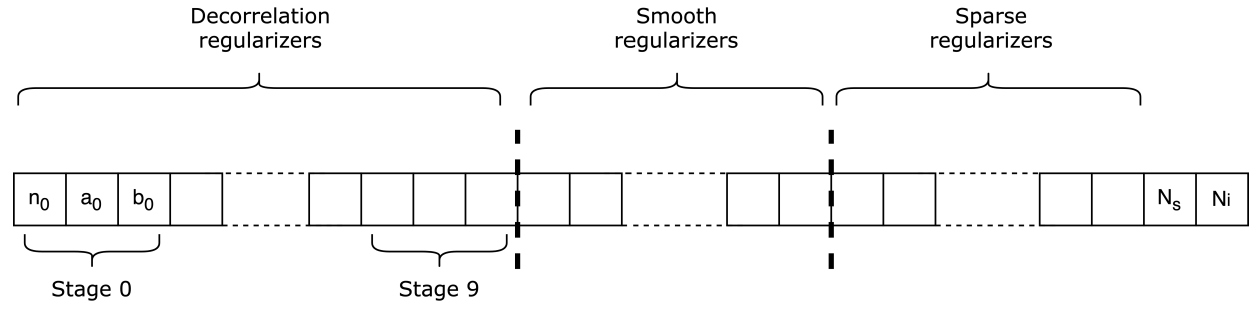}
  \caption{Vectorization scheme for surrogate modeling in graph-based approach.}
  \label{fig:vect}
  
\end{figure}

For the fixed-size solution, surrogate implementation is native \cite{10.1093/jigpal/jzac019}. In case of graph-based pipeline, firstly, we allocate a place for all 3 possible types of regularizers. Secondly, for every regularizer type, we allocate a place for the maximum number of stages for this regularizer (10). Finally, for each possible stage, we allocate a place for the number of iterations $n_i$ and a place for the parameters related to this regularizer (2 for all types of the considered regularizers: $a_i$, $b_i$). Figure~\ref{fig:vect} illustrates the proposed vectorization scheme. Pairs of vectors and corresponding fitness values are used for training the surrogate ML model.

Available in AutoTM surrogate models include Gaussian Process and Random Forest variants.

\begin{figure}[h]
  \centering
  \includegraphics[width=0.4\linewidth]{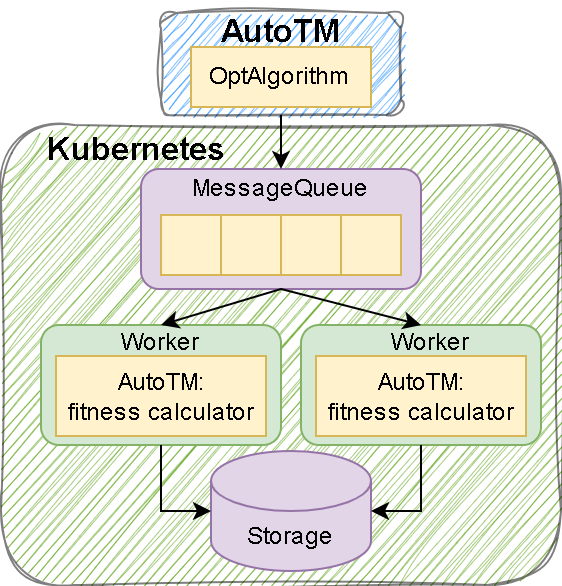}
  \caption{Distributed mode schema.}
  \label{fig:distr_mode}
\end{figure}

\begin{figure*}[h!]
    \centering
    \subfigure[]{\includegraphics[width=0.4\textwidth]{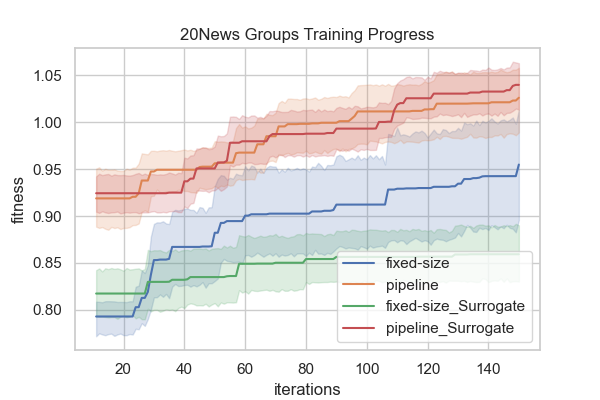}} 
    \subfigure[]{\includegraphics[width=0.4\textwidth]{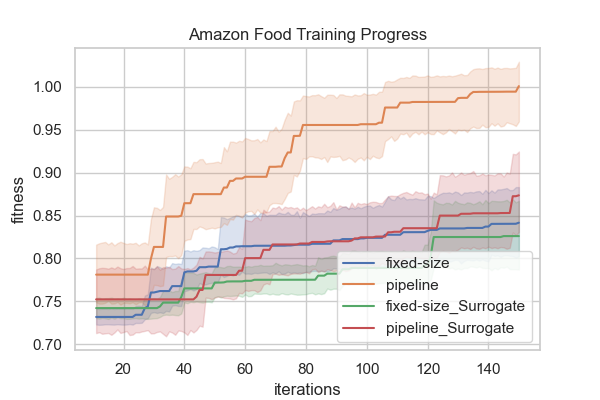}} 
    \subfigure[]{\includegraphics[width=0.4\textwidth]{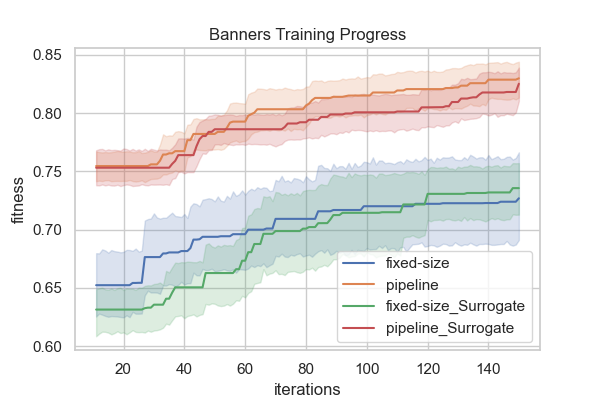}}
    \subfigure[]{\includegraphics[width=0.4\textwidth]{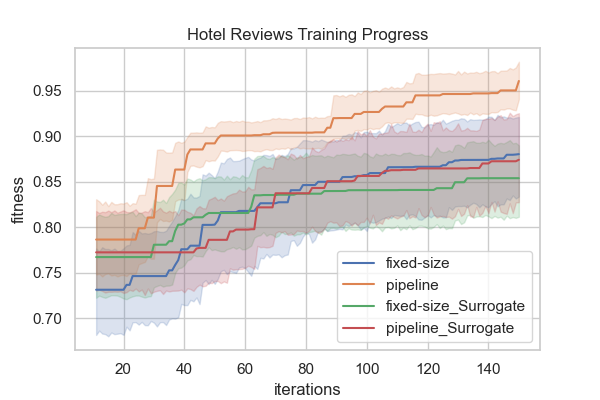}}
    \subfigure[]{\includegraphics[width=0.4\textwidth]{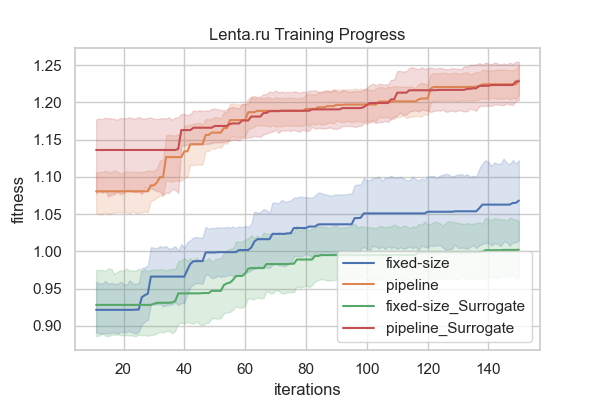}}
    \caption{Average fitness values with 90\% confidence interval for 5 datasets ((a) 20 Newsgroups (b) Amazon food (c) Banners (d) Hotel reviews (e) Lenta.ru) by the number of used iterations with the usage of a surrogate model.}
    \label{fig:panel_surrogate}
\end{figure*}

\subsection{Distributed mode}

The framework supports distributed fitness computations for evaluation of newly created individuals. The mode relies on a set of independent workers deployed on the cluster and a message queue to spread the workload (e.g. individuals to be estimated) among the workers. Each worker has access to one or more dataset being processed at the time and trains a topic model according to the parameters stored in an individual. The main process is able to use it by means of a celery app, which provides a map-like interface to submit multiple individuals for evaluation at once (fig~\ref{fig:distr_mode}). For convenience, we provide a helm chart to deploy workers on Kubernetes.

\section{Framework Performance}

\subsection{Datasets overview}
The datasets we used in our experiments are identical to those used in the original AutoTM study [21]. This choice was made to directly align
our efforts with the previous study, facilitating a more accurate comparison of
results. These datasets vary considerably in terms of size and structure, thereby
encompassing a broad array of potential scenarios: 1) \textbf{20 Newsgroups} dataset \cite{LANG1995331} consists of around 180,000 posts from newsgroups that cover 20 different topics. Given the variety of topics, it is acknowledged as a complex real-world dataset that poses a considerable difficulty for text analysis and concept learning applications; 2) \textbf{Amazon Fine Food Reviews} corpus \cite{10.1145/2488388.2488466} includes a range of food options that can be found on Amazon; 3) \textbf{Banners Pages} Dataset \cite{nevezhin-etal-2020-topic} includes pages that banners direct to, which were gathered from the internet. It comprises 400,000 advertisement pages that cover a
diverse range of subjects; 4) \textbf{Datainfini’s Hotel Reviews} dataset\footnote{https://www.kaggle.com/datasets/datafiniti/hotel-reviews} includes 34,399 guest reviews of 1,000 hotels, highlighting various aspects of service quality; 5) \textbf{Lenta.ru} Dataset — this Russian-language news corpus\footnote{https://github.\%20com/yutkin/Lenta.Ru-News-Dataset} contains over 600,000 news articles collected from the years 1999 to 2019.

To evaluate topic model quality, we used dataset\footnote{https://www.kaggle.com/datasets/marykh/marked-up-topic-quality} with manually evaluated topics on three different datasets \cite{9953874}. 

\subsection{Pipelines comparison}

According to results in \cite{Khodorchenko2023-hy} for AutoTM can be compared in quality with existing topic modeling frameworks, thus we are going to show the improvement of a newly proposed pipeline that combines graph-based  over the previous one.

\begin{figure}[h]
  \centering
  \includegraphics[width=0.7\linewidth]{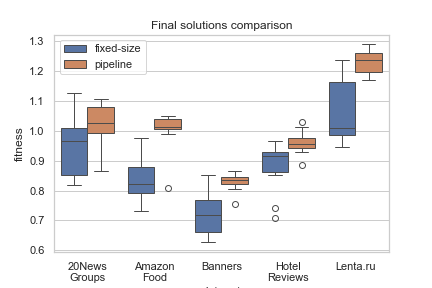}
  \caption{Comparison between two different ways of chromosome representation.}
  \label{fig:optimus}
\end{figure}

The method proposed in this study showed (see Fig.~\ref{fig:optimus}) a significant enhancement in terms of quality as follows: an improvement of +7.4\% was observed for the 20New Groups dataset, +18.9\% for Amazon Food, +14.2\% for Banners,
+9.2\% for Hotel Reviews, and +15.1\% for Lenta.ru. On the whole, the proposed approach demonstrated an average improvement of 13.0\% in fitness compared with the original solution.

We also show the results for different pipelines with and without surrogate models (fig~\ref{fig:panel_surrogate})
The proposed surrogate solution has been found to notably improve
computational efficiency by 10\%, while still delivering high-quality results. By
implementing the suggested solution with a surrogate model, the average quality loss is minimal, with only a 4\% decrease compared to running the system
without using a surrogate.

\subsection{Quality metrics performance}

Table~\ref{fig:table_tm_quality} show the performance of existing metrics in terms of alignment with human judgement. Headers with "25", e.g., "Coherence 25" indicate the amount of tokens with the highest probability that were used for scoring. It is evident that gpt-based evaluation provides the best results for two datasets, but struggles with the third one. It is an indicator of a potential need in defining a more specific metric for the dataset at hand.

\begin{table}[h!]
\centering
\begin{tabular}{|l|l|l|l|l|l|}
\hline
                                                      & Gpt4o & Default fitness & Coherence 25 & NPMI 25 & SwitchP \\ \hline
20ng                                                  & \textbf{0.78}  & 0.71                                                        & 0.62                                              & 0.74                                              & -0.65   \\ \hline
amazon food & \textbf{0.66}  & 0.2                                                         & 0.36                                              & 0.18                                              & -0.34   \\ \hline
lenta                                                 & 0.64  & 0.86                                                        & \textbf{0.92}                                              & 0.63                                              & 0.04    \\ \hline
\end{tabular}
\caption{Correlation with human judgement for a range of evaluation metrics, with the highest correlations}
\label{fig:table_tm_quality}
\end{table}

\section{Conclusion}
We presented an AutoTM 2.0 framework for optimizing additively regularized topic models. The framework allows to effortlessly train and utilize topic models for datasets with various characteristics without requiring extensive expertise in ARTM hyperparameters. Flexible pipelines, heuristics and LLM based quality metrics, surrogate-assistance and distributed mode are all the improvements freshly introduced to the framework which enables to achieve significantly better results up to 13.0\% on average in comparison with the previous version.



\bibliographystyle{unsrt}  
\bibliography{references}  


\end{document}